\documentclass[10pt,twocolumn,letterpaper]{article}

\usepackage[pagenumbers]{cvpr}

\usepackage[dvipsnames]{xcolor}

\definecolor{cvprblue}{rgb}{0.21,0.49,0.74}
\usepackage[pagebackref,breaklinks,colorlinks,citecolor=cvprblue]{hyperref}
\usepackage{color}
\usepackage{colortbl}
\usepackage{caption}
\usepackage{array}
\usepackage{multirow}
\usepackage{booktabs}
\usepackage{pifont}
\usepackage{amssymb}
\usepackage{amsmath}
\usepackage{bm}
\usepackage{tikz}
\definecolor{Gray}{gray}{0.93}
\usepackage{listings}
\usepackage{newfloat}
\usepackage{footnote}
\title{PVLR: Prompt-driven Visual-Linguistic Representation Learning \\for Multi-Label Image Recognition}

\author{
\textbf{Hao Tan}\textsuperscript{\rm 1,\rm 2}\footnote[2]{}, 
\textbf{Zichang Tan}\textsuperscript{\rm 3}\footnote[2]{}, 
\textbf{Jun Li}\textsuperscript{\rm 1,\rm 2}, 
\textbf{Jun Wan}\textsuperscript{\rm 1,\rm2}\thanks{Corresponding author. $^{\dagger}$  \text{Equal contribution.}}, 
\textbf{Zhen Lei}\textsuperscript{\rm 1,\rm2,\rm4}\\
\textsuperscript{\rm 1}MAIS, Institute of Automation, Chinese Academy of Sciences, Beijing, China\\
\textsuperscript{\rm 2}School of Artificial Intelligence, University of Chinese Academy of Sciences, Beijing, China\\
\textsuperscript{\rm 3}Department of Computer Vision Technology (VIS), Baidu Inc., Beijing, China\\
\textsuperscript{\rm 4}CAIR, HKISI, Chinese Academy of Sciences, Hong Kong, China\\
{\tt\small \{tanhao2023, lijun2021, jun.wan, zhen.lei\}@ia.ac.cn, tanzichang@baidu.com}
}

\begin{document}
\maketitle
\begin{abstract}
Multi-label image recognition is a fundamental task in computer vision. 
Recently, vision-language models have made notable advancements in this area. 
However, previous methods often failed to effectively leverage the rich knowledge within language models and instead incorporated label semantics into visual features in a unidirectional manner. 
In this paper, we propose a \textbf{P}rompt-driven \textbf{V}isual-\textbf{L}inguistic \textbf{R}epresentation Learning (PVLR) framework to better leverage the capabilities of the linguistic modality. 
In PVLR, we first introduce a dual-prompting strategy comprising Knowledge-Aware Prompting (KAP) and Context-Aware Prompting (CAP). 
KAP utilizes fixed prompts to capture the intrinsic semantic knowledge and relationships across all labels, while CAP employs learnable prompts to capture context-aware label semantics and relationships. 
Later, we propose an Interaction and Fusion Module (IFM) to interact and fuse the representations obtained from KAP and CAP. 
In contrast to the unidirectional fusion in previous works, we introduce a Dual-Modal Attention (DMA) that enables bidirectional interaction between textual and visual features, yielding context-aware label representations and semantic-related visual representations, which are subsequently used to calculate similarities and generate final predictions for all labels. 
Extensive experiments on three popular datasets including MS-COCO, Pascal VOC 2007, and NUS-WIDE demonstrate the superiority of PVLR.
\end{abstract}  

\section{Introduction}
\label{sec:intro}
\begin{figure}[t]
    \centering
    \includegraphics[width=0.97\linewidth]{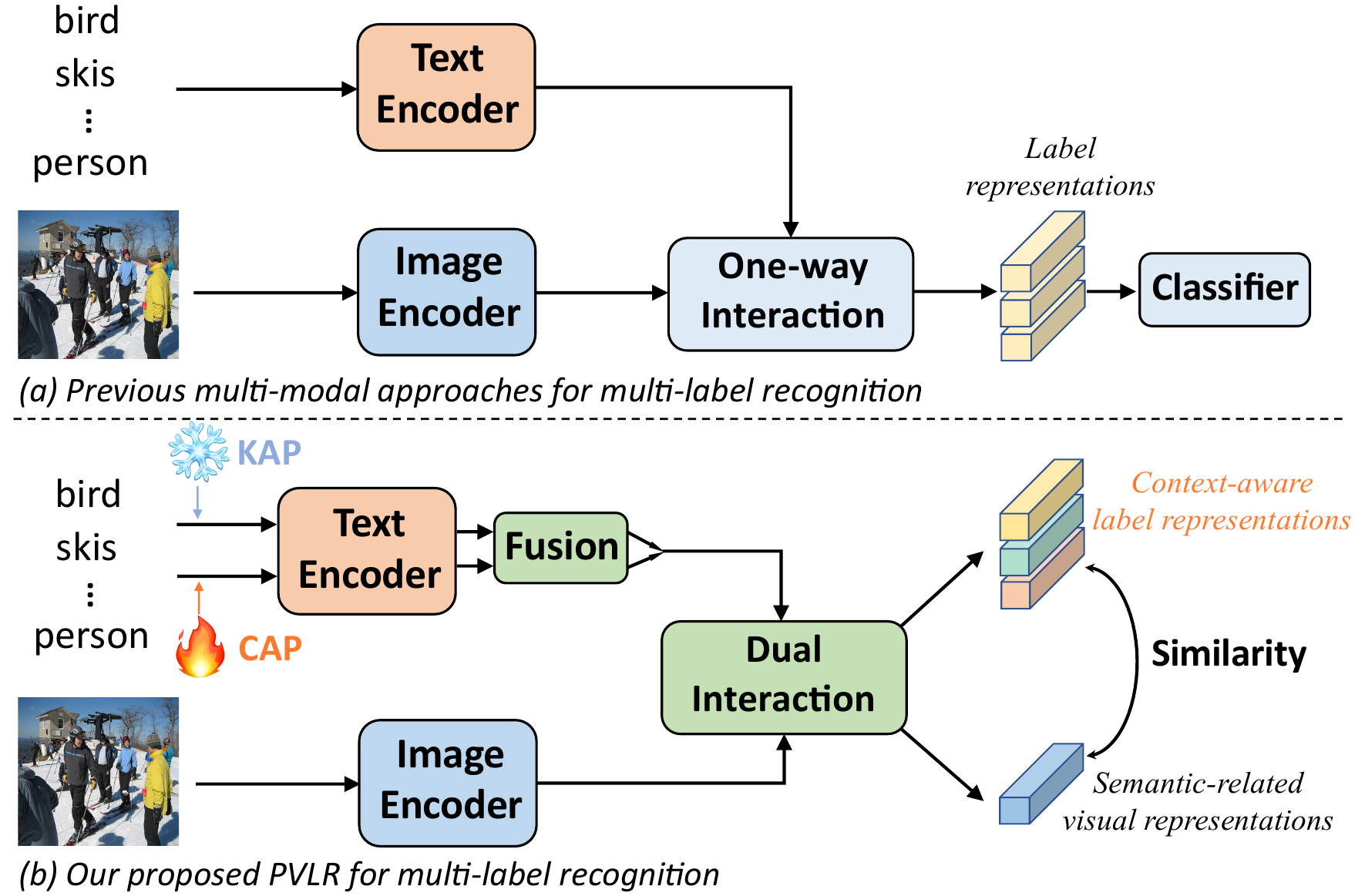}
    \caption{\textbf{Paradigm comparison.} (a) Previous methods~\cite{chen2019learning, you2020cross, wang2020multi, zhu2022two, zhu2023scene} simply adopt category names to extract text features and they apply a one-way interaction to yield label representations. Then $C$ classifiers are trained to perform recognition. (b) Our method treats the two modalities of equal importance, where we propose dual prompting to extract text features and perform dual interaction to generate context-aware label representations and semantic-related visual representations.}
	\label{fig:intro} 
\end{figure}
Multi-label image classification~\cite{wang2016cnn, wang2017multi, chen2018order, chen2019learning, wang2020multi, ridnik2021asymmetric, chen2019multi, you2020cross, yazici2020orderless, ye2020attention, zhao2021m3tr, liu2021query2label, lanchantin2021general, zhu2022two, ridnik2023ml, zhu2023scene, li2023patchct, guo2023texts}
is a fundamental task in the field of computer vision,
where multiple labels are supposed to be recognized in a single image.
This ability to capture the diversity of visual content is paramount in applications like image tagging~\cite{huang2023tag2text, mamat2023enhancing}, human attribute recognition~\cite{tan2019attention, tan2020relation}, and recommendation systems~\cite{yang2015pinterest, jain2016extreme}.

With the rise of vision-language pre-training~\cite{radford2021learning, jia2021scaling}, 
many approaches~\cite{chen2019learning, you2020cross, wang2020multi, zhao2021m3tr, zhu2022two, zhu2023scene, li2023patchct} have leveraged linguistic modality to mitigate the lack of semantic information from a single visual input.
Typically, these approaches involve extracting the inherent semantic knowledge within the language model and employing it as supplementary information to assist the visual model in learning better label representations.
Due to the extensive semantic knowledge embedded in language models, these methods have made some improvements in multi-label recognition. 

However, the application of vision-language modeling in the domain of multi-label image recognition is still in its infancy. 
One notable limitation is the underutilization of linguistic modality.
As summarized in Figure~\ref{fig:intro}, such limitation often manifests in two aspects:
\textbf{1)} Previous methods~\cite{chen2019learning, you2020cross, wang2020multi, zhu2022two, zhu2023scene} simply take the static category names as the inputs of language models (e.g. BERT~\cite{devlin2018bert}), 
which lack sufficient knowledge acquisition and interaction for these powerful models since they are typically pre-trained with complete context (e.g., complete sentences).
Such complete context can facilitate the capture of task-relevant semantic information.
\textbf{2)} The linguistic modality is merely adopted as a supplement of semantic information through a one-way interaction with visual features, which hinders the role of linguistic modality. 
Thus, $C$ additional classifiers ($C$ is the number of candidate labels) are needed to learn the category centers, which remain static during inference.

In order to address the aforementioned problems, we propose a \textbf{P}rompt-driven \textbf{V}isual-\textbf{L}inguistic \textbf{R}epresentation Learning (PVLR) framework for multi-label image recognition.
To tackle the first issue, we propose a dual-prompting strategy, namely Knowledge-Aware Prompting (KAP) and Context-Aware Prompting (CAP).
Specifically, the recent success of zero-shot predictions~\cite{radford2021learning, yu2023zero} by utilizing hand-crafted templates such as ``\textit{a photo of a [CLS].}'' motivates us to introduce textual prompts in KAP to facilitate the extraction of general knowledge from language models.
However, the KAP remains static and is not informative enough for the vastly changing visual context. 
Therefore, inspired by recent advancements in prompt learning~\cite {zhou2022learning, zhou2022conditional}, we further introduce learnable prompts in CAP to facilitate the learning of task-relevant semantics, and adaptively incorporate the visual information into label semantics.
To deeply aggregate the information learned by KAP and CAP, we further propose an Interaction and Fusion Module (IFM).    
The motivation behind the proposed IFM stems from the observation that we humans, when assessing the presence of objects in our surroundings, not only rely on visual information but also employ our prior knowledge to make educated guesses about objects that are partially occluded or viewed from a distant perspective. 
In this way, we first integrate the knowledge-guided attention map and context-aware attention map generated by KAP and CAP.
Then we perform a channel interaction between the label embeddings extracted from KAP and CAP to produce the fused representations for all labels based on the aggregated attention map.

To overcome the second problem,
we propose a Dual-Modal Attention (DMA) module to treat visual and linguistic modalities of equal importance, which effectively harnesses the advantages of the linguistic modality and captures impactful visual-linguistic representations. Different from existing methods~\cite{chen2019learning, you2020cross, wang2020multi, zhu2022two, zhu2023scene} where textual information is only unidirectionally integrated with visual information,
our DMA allows a bidirectional interaction between the two modalities. 
To achieve this, the proposed DMA consists of two attention modules, i.e., visual-to-semantic attention and semantic-to-visual attention.
Specifically, the visual-to-semantic attention integrates the visual information into the extracted label embeddings, yielding \textit{context-aware label representations}, 
while the semantic-to-visual attention integrates the extracted label semantics into visual features, resulting in \textit{semantic-related visual representations}. 
Unlike previous methods~\cite{chen2019learning, you2020cross, wang2020multi, zhu2022two, zhu2023scene} that employ fixed classification weights,
our method predicts the scores based on the similarity between these two representations, which achieves \textit{input-adaptive category centers}, largely enhancing the model's generalization capability. 
To sum up, the main contributions of this work include:
\begin{itemize}
    \item We propose PVLR, a novel visual-linguistic representation learning framework for multi-label image recognition, which achieves state-of-the-art performance on three widely used benchmarks.
    \item We emphasize that appropriate prompting can facilitate knowledge extraction from the language model, where we propose KAP and CAP to extract general knowledge and task-relevant semantics respectively. An IFM is further proposed to aggregate their information, which yields powerful prompt-driven label representations.
    \item Rather than unidirectionally leveraging linguistic information, we propose to perform bidirectional interactions between visual and linguistic modalities, where we generate context-aware label representations and semantic-related visual representations concurrently through DMA, and achieve input-adaptive category centers.
\end{itemize}

\begin{figure*}[t]
    \centering
    \includegraphics[width=0.99\linewidth]{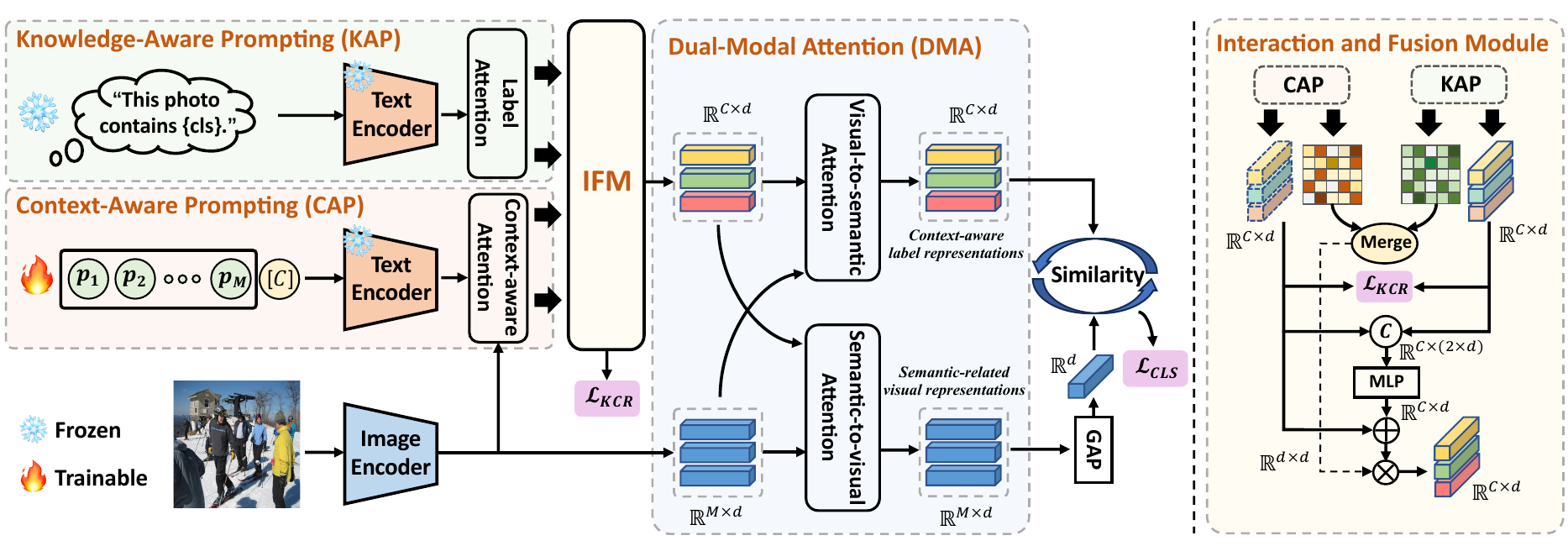}
 \caption{\textbf{Overview of the proposed PVLR framework.} 
 \textbf{1)} KAP adopts fixed prompts to facilitate the extraction of semantic knowledge from the language model.
 \textbf{2)} CAP utilizes learnable prompts to integrate context clues into label representations.
 \textbf{3)} IFM deeply aggregates the information learned by KAP and CAP through channel interaction. And $\mathcal{L}_{KCR}$ is measured to enhance the generalization.
 \textbf{4)} DMA performs bidirectional attention to generate context-aware label representations and semantic-related visual representations, respectively.
 The context-aware label representations are then regarded as the classification weights and the prediction is based on the similarity between these two representations, which achieves input-adaptive category centers.
 }
	\label{fig:framework}
\end{figure*}
\section{Related Work}
\label{sec:related}
\subsection{Multi-Label Classification}
Multi-label recognition serves as a fundamental task in the field of computer vision.
Early approaches~\cite{wang2016cnn, yazici2020orderless, chen2018order, wang2017multi, ye2020attention, lanchantin2021general} consider a single visual modality as input and focus on the modeling of label co-occurrence.
Some of them rely on the Recurrent Neural Network (RNN)~\cite{hochreiter1997long} and graph-based models~\cite{kipf2016semi}.
For example, Wang et al.~\cite{wang2016cnn} explored the semantic correlation among labels by cascading the RNN to the feature extractor.
Wang et al.~\cite{wang2017multi} utilized the LSTM to capture the dependencies among semantic regions.
Ye et al.~\cite{ye2020attention} proposed a dynamic graph convolutional network (GCN) to model the content-aware relations for co-occurred categories.
Further works turn to visual attention for implicit relation mining.
For instance, Lanchantin et al.~\cite{lanchantin2021general} utilized a transformer encoder to explore the correlations among visual features and labels.
Zhu et al.~\cite{zhu2017learning} introduced self-attention to capture spatial relationships and then regularized the predictions.
Most recent works further enhance the performance from the perspective of knowledge distillation~\cite{rajeswar2022multi, yang2023multi} and loss improvements~\cite{kobayashi2023two}.
However, uni-modal methods lack label-related semantic information, which limits their generalization ability.
With the rise of language models such as BERT~\cite{devlin2018bert}, many approaches~\cite{chen2019learning, chen2019multi, wang2020multi, you2020cross, liu2021query2label, zhu2022two, zhao2021m3tr, zhu2023scene, li2023patchct} turn to leveraging linguistic modality to complement the semantic information.
Based on the extracted representations for candidate labels, these methods further focus on the interactions of different modalities.
You et al.~\cite{you2020cross} proposed a cross-modality attention module to aggregate visual features and label embeddings, while $C$ classifiers are learned for classification.
Similarly, Zhu et al.~\cite{zhu2022two} constructed a two-stream transformer network to explore the textual-visual interactions and an MLP is trained for recognition.
Wang et al.~\cite{wang2020multi} superimposed multiple layers of graph to incrementally inject label semantics to feature learning.
Chen et al.~\cite{chen2019learning} and Zhu et al.~\cite{zhu2023scene} both inserted semantic information to visual features unidirectionally through a low-rank bilinear pooling, while~\cite{zhu2023scene} further consider the scene-conditioned label co-occurrence.

Different from them, we explore the mutual interactions between visual and linguistic modality, and further map context-aware label representations into category centers.
Although similar approach has appeared in~\cite{chen2019multi}, the learning of category centers in their method is agnostic to visual clues.
In contrast, the category centers in our work are adaptive to input context, which enhances the generalization.

\subsection{Vision-Language Models}
Large-scale vision-language pre-training has emerged as a powerful paradigm for a wide range of visual tasks~\cite{luo2022clip4clip, guo2023texts, zhu2022pointclip}.
With a contrastive-based pre-training approach, vision-language models (VLMs) such as CLIP~\cite{radford2021learning} and ALIGN~\cite{jia2021scaling} learn a joint representation for visual and linguistic modalities, showing an encouraging ability for efficient transfer learning and zero-shot predictions.
More recently, there has been a growing interest in bridging pre-trained LLM and vision foundation models to build VLM.
Flamingo~\cite{alayrac2022flamingo} introduced gated attention for modality interactions and achieved promising few-shot abilities.
BLIP-2~\cite{li2023blip} achieved this by training an additional Q-former.
While MiniGPT-4~\cite{zhu2023minigpt} and LLaVA~\cite{liu2023visual} attained impressive multi-modal abilities by only adding a linear projection.

Although previous works have made use of these powerful VLMs, most of them lack sufficient knowledge acquisition from the language model.
In contrast, we propose a dual prompting strategy to facilitate the extraction of general knowledge and task-relevant label semantics.

\section{Proposed Method}
\label{sec:method}
The overall pipeline of our PVLR framework is shown in Figure~\ref{fig:framework}. 
In this section, we first define some basic notations in Section~\ref{sec:pre}, and then introduce the proposed KAP (Section~\ref{sec:kap}), CAP (Section~\ref{sec:cap}), IFM (Section~\ref{sec:ifm}) and DMA (Section~\ref{sec:dma}) in detail.

\subsection{Preliminary}
\label{sec:pre}
\textbf{Notations.} 
For multi-label image recognition,
assume the input image $\boldsymbol{I}\in \mathbb{R}^{H\times W \times 3}$ is labeled with $C$ candidate categories, where $\boldsymbol{y}\in \mathbb{R}^{C}$ represents the multi-hot label vector and $\boldsymbol{y}_{j}=1$ means the input image contains the $j^{th}$ label and vice versa. 
For the input image $\boldsymbol{I}$,
we employ an image encoder (e.g., ResNet~\cite{he2016deep} or ViT~\cite{dosovitskiy2020image})
to extract visual features$\boldsymbol{X} \in \mathbb{R}^{M \times d}$,
where $M$ indicates the number of pixels or patches, and $d$ is the feature dimension.

\noindent\textbf{Attention mechanism.}
Transformer has achieved significant success in visual tasks~\cite{dosovitskiy2020image, chen2021crossvit, arnab2021vivit}, 
particularly due to its well-designed attention mechanism, which exhibits strong capability in relation modeling.
Typically, there are two kinds of attention mechanisms in the transformer, i.e., self-attention and cross-attention.
For self-attention, it models the relations among the elements within an input sequence $\boldsymbol{E} \in \mathbb{R}^{M\times d}$ ($M$ is the number of vectors and $d$ denotes the dimension of the features), which is formulated as:
\begin{equation}
\begin{aligned}
    &\text{Self-Attn}(\boldsymbol{E}) = \text{softmax}( \frac{\boldsymbol{Q}\boldsymbol{K}^{\mathsf{T}}}{\sqrt{d}})\boldsymbol{V}, \\
    &\text{where}\ \boldsymbol{Q}=\boldsymbol{E}\boldsymbol{W}_Q, \boldsymbol{K}=\boldsymbol{E}\boldsymbol{W}_K, \boldsymbol{V}=\boldsymbol{E}\boldsymbol{W}_V,
\end{aligned}
\end{equation}
where $\boldsymbol{W}_Q$, $\boldsymbol{W}_K$ and $\boldsymbol{W}_V$ are learnable weights. 
We take $\boldsymbol{\mathcal{M}} = \text{softmax}( \frac{\boldsymbol{Q}\boldsymbol{K}^{\mathsf{T}}}{\sqrt{d}})$ to denote the attention map that captures the pair-wise relations of vectors in $\boldsymbol{E}$.
In contrast, cross-attention takes different sources as input and is good at capturing cross-domain interactions.
Suppose the inputs are denoted as $\boldsymbol{E}$ and $\boldsymbol{Z}$, the process is formulated as:
\begin{equation}
\begin{aligned}
    &\text{Cross-Attn}(\boldsymbol{E}, \boldsymbol{Z}) = \text{softmax}( \frac{\boldsymbol{Q}\boldsymbol{K}^{\mathsf{T}}}{\sqrt{d}})\boldsymbol{V}, \\
    &\text{where}\ \boldsymbol{Q}=\boldsymbol{E}\boldsymbol{W}_Q, \boldsymbol{K}=\boldsymbol{Z}\boldsymbol{W}_K, \boldsymbol{V}=\boldsymbol{Z}\boldsymbol{W}_V.
\end{aligned}
\end{equation}

\subsection{Knowledge-Aware Prompting}
\label{sec:kap}
Large VLMs~\cite{radford2021learning, jia2021scaling} typically encompass a wealth of semantic knowledge since the pre-training on large-scale image-text pairs. 
Therefore, by setting appropriate textual inputs for each label, 
the embeddings extracted by the language model will contain the underlying semantic relations between different labels.

\noindent\textbf{Hard prompts.}
Specifically, we take the hand-crafted templates (i.e., hard prompts) ``\textit{This photo contains [CLS].}'' as the inputs for all $C$ labels.
Then the text encoder is adopted to extract $C$ label embeddings, which are denoted as $\boldsymbol{T}^{hard}=\{\boldsymbol{t}_1^{hard},...,\boldsymbol{t}_{C}^{hard}\}\in \mathbb{R}^{C\times d}$, 
where $d$ is the hidden dimension of CLIP. 

\noindent\textbf{Label attention.}
Moreover, modeling relations among different labels helps the discovery of the label co-occurrence.
Therefore, we perform a label attention on the label embeddings to capture the knowledge-guided relations,
which is formally written as:
\begin{equation}
    \boldsymbol{T}^{ka}, \boldsymbol{\mathcal{M}}^{ka} = \text{Self-Attn}(\boldsymbol{T}^{hard}),
\end{equation}
where $\boldsymbol{T}^{ka}\in \mathbb{R}^{C\times d}$ is the relation-enhanced label embeddings.
$\boldsymbol{\mathcal{M}}^{ka}\in \mathbb{R}^{C\times C}$ denotes the knowledge-guided attention map, where $\boldsymbol{\mathcal{M}}^{ka}_{ij}$ depicts the relation between $\boldsymbol{t}_i^{hard}$ and $\boldsymbol{t}_j^{hard}$, indicating the underlying co-occurrence probability of the $i^{th}$ and the $j^{th}$ label.

\subsection{Context-Aware Prompting}
\label{sec:cap}
As mentioned above, KAP could extract rich semantic knowledge and relations.
However, such information is static and independent of the input image.
To address this, we further propose a Context-Aware Prompting (CAP) module to capture context-aware label relations. 
Compared to KAP, there are mainly two differences in CAP:
1) CAP adopts learnable tokens (i.e., soft prompts) to prompt the language model, which facilitates the learning of downstream semantics, allowing for fine-grained relation exploration.
2) CAP employs a context-aware attention to adaptively aggregate context information into label representations and models the context-aware label relation.

\noindent\textbf{Soft prompts.} 
Inspired by~\cite{zhou2022learning, zhou2022conditional}, we prepend $L$ prompt tokens to each label and yield ``$[\boldsymbol{p}_1][\boldsymbol{p}_2]...[\boldsymbol{p}_L][\boldsymbol{s}_j]$'', where $\boldsymbol{p}_l\in \mathbb{R}^{d}$ is learnable to adapt the task and $\boldsymbol{s}_j$ is the word embedding of the $j^{th}$ label name.
Then the sequences are fed into text encoder to extract $C$ label embeddings, which are denoted as $\boldsymbol{T}^{soft}=\{\boldsymbol{t}_1^{soft},...,\boldsymbol{t}_{C}^{soft}\}\in \mathbb{R}^{C\times d}$.

\noindent\textbf{Context-aware attention.}
To incorporate fine-grained context clues into static label semantics, we condition the label embeddings on visual features, which is formulated as:
\begin{equation}
\begin{aligned}
    &\boldsymbol{T}^{'soft} = \text{Cross-Attn}(\boldsymbol{T}^{soft}, \boldsymbol{X}), \\
    &\boldsymbol{T}^{ca}, \boldsymbol{\mathcal{M}}^{ca} = \text{Self-Attn}(\boldsymbol{T}^{'soft}),
\end{aligned}
\end{equation}
where $\boldsymbol{T}^{'soft}$ captures the interaction between each label representation and all spatial regions.
$\boldsymbol{T}^{ca}$ is the context-aware label representations and $\boldsymbol{\mathcal{M}}^{ca}$ represents the context-aware attention map.

\subsection{Interaction and Fusion Module}
\label{sec:ifm}
To fully leverage the strengths of the dual-prompting, we propose an Interaction and Fusion Module (IFM).
This consists of two aspects, i.e., the interaction between prompts and the aggregation of relations.

\noindent\textbf{Prompts interaction.}
To deeply aggregate the information learned by KAP and CAP, we perform an interaction between $\boldsymbol{T}^{ka}$ and $\boldsymbol{T}^{ca}$.
Specifically, we propose a channel interaction to continually inject general knowledge to $\boldsymbol{T}^{ca}$,
which is formulated as:
\begin{equation}
    \boldsymbol{T}^{ca} = \boldsymbol{T}^{ca} + \text{MLP}([\boldsymbol{T}^{ka}, \boldsymbol{T}^{ca}]),
\end{equation}
where $\text{MLP}(\cdot)$ denotes a Multi-Layer Perceptron and $[\cdot]$ denotes the concatenation operation.
For simplicity, we reuse the notation $\boldsymbol{T}^{ca}$ for the modulated embedding.
Moreover, soft prompting is potential to overfit the seen data and forget the general knowledge~\cite{bulat2023lasp, yao2023visual}.
Therefore, we further introduce a knowledge-to-context regularization (KCR) loss to enhance the generalization ability, which is formulated as:
\begin{equation}
    \mathcal{L}_{KCR} = \frac{1}{C} \sum_{j=1}^{C}(1 - \frac{\boldsymbol{t}_j^{ka}(\boldsymbol{t}_j^{ca})^{\mathsf{T}}}{||\boldsymbol{t}_j^{ka}|| ||\boldsymbol{t}_j^{ca}||}).
\end{equation}
\noindent\textbf{Relation aggregation.}
Besides, the relation among labels should consider both general knowledge and practical contexts.
Therefore, we propose to aggregate knowledge-guided attention map $\boldsymbol{\mathcal{M}}^{ka}$ and context-aware attention map $\boldsymbol{\mathcal{M}}^{ca}$ through a re-weighting scheme.
The aggregated map is then adopted to enhance the label representations:
\begin{equation}
    \boldsymbol{T} = (\alpha \boldsymbol{\mathcal{M}}^{ka} + (1-\alpha) \boldsymbol{\mathcal{M}}^{ca})\boldsymbol{T}^{ca},
\end{equation}
where $\alpha$ is set to be learnable and $\boldsymbol{T}\in \mathbb{R}^{C\times d}$ denotes the relation-enhanced label representations.

\subsection{Dual-Modal Attention}
\label{sec:dma}
So far, we have captured the prompt-driven label representations, while the mutual interaction between visual and linguistic modalities is still underexplored.
Inspired by~\cite{nam2017dual, yuan2021florence}, we propose to interact visual features and label representations through a dual attention mechanism, which consists of two attention modules, i.e., visual-to-semantic attention and semantic-to-visual attention modules.
 
\noindent\textbf{Visual-to-semantic attention.} 
To integrate visual information into label representations, we take label representations $\boldsymbol{T}$ as query and perform a cross-attention with visual features, which is formally written as: 
\begin{equation}
    \boldsymbol{T}^{v\rightarrow s} = \text{Cross-Attn}(\boldsymbol{T}, \boldsymbol{X}).
\end{equation}

\noindent\textbf{Semantic-to-visual attention.} 
To inject semantic information into visual representations, we take visual features $\boldsymbol{X}$ as query and perform a cross-attention with label representations, which is formulated as:
\begin{equation}
    \boldsymbol{X}^{s\rightarrow v} = \text{Cross-Attn}(\boldsymbol{X}, \boldsymbol{T}).
    \label{eq_s2v_att}
\end{equation}
Then the generated visual features are averaged through a global average pooling (GAP),
and we omit the process in the Eq.~\ref{eq_s2v_att} for simplicity.
The acquired label embeddings $\boldsymbol{T}^{v\rightarrow s}\in \mathbb{R}^{C\times d}$ are aware of visual context and serve as powerful representations for candidate labels.
The resulted visual feature $\boldsymbol{X}^{s\rightarrow v}\in \mathbb{R}^{d}$ is closely related to the label semantic and is robust to perform final recognition.

Different from previous works~\cite{chen2019learning, you2020cross, wang2020multi, zhu2022two, zhu2023scene} that employ fixed classification weights (e.g., linear layers) for recognition, we regard each label representation as the center of the corresponding category,
which is an input-adaptive manner and helps to enhance the model's generalization capability.
To be specific, the presence probability of the $j^{th}$ label is predicted through measuring the similarity between visual representation $\boldsymbol{X}^{s\rightarrow v}$ and the $j^{th}$ label representation $\boldsymbol{T}^{v\rightarrow s}_j$:
\begin{equation}
    \boldsymbol{p}_j = \sigma(\boldsymbol{X}^{s\rightarrow v} (\boldsymbol{T}^{v\rightarrow s}_j)^{\mathsf{T}}),
    \label{eq_p}
\end{equation}
where $\sigma(\cdot)$ is the sigmoid function to map the predicted logit into a probability.

\subsection{Training Objective}
\label{sec:objective}
Based on final predictions in Eq.~\ref{eq_p}, the Asymmetric Loss~\cite{ridnik2021asymmetric} is employed for multi-label classification:
\begin{equation}
    \left.\mathcal{L}_{CLS}=\frac{1}{C}\sum_{j=1}^{C}\left\{\begin{aligned}(1-\boldsymbol{p}_{j})^{\gamma^+}\log \boldsymbol{p}_{j},\quad \boldsymbol{y}_{j}&=1,\\ \boldsymbol{p}_{j}^{\gamma^-}\log(1-\boldsymbol{p}_{j}),\quad \boldsymbol{y}_{j}&=0,
\end{aligned}\right.\right.
\label{eq:gamma}
\end{equation}
where $\gamma^+$ and $\gamma^-$ are asymmetric focusing parameters for positive and negative samples, respectively.

Together with the hard-to-soft regularization loss, the final objective is defined as:
\begin{equation}
    \mathcal{L} = \mathcal{L}_{CLS} + \lambda \mathcal{L}_{KCR},
\label{eq:loss}
\end{equation}
where $\lambda$ is a hyper-parameter to make a trade-off between the two losses.

\section{Experiments}
\label{sec:exp}
\subsection{Datasets and Metrics}

\textbf{MS-COCO.}
Microsoft COCO~\cite{lin2014microsoft} is the most widely used dataset to evaluate the task of multi-label classification.
It contains 123,287 images in total with 80 categories, where each image has about 2.9 labeled objects.
We train the model with training set which contains 82,783 images and evaluate on the test set with 40,504 samples.

\noindent\textbf{PASCAL VOC 2007.}
VOC 2007~\cite{everingham2010pascal} is a popular benchmark for multi-label recognition. It includes 9,963 images in total with 20 distinct object classes.
Each image is annotated with 1.4 labels in average.
Following~\cite{chen2019multi}, we train the model on the \textit{train-val} set which contains 5,011 images and evaluate on the \textit{test set} with 4,952 images.

\noindent\textbf{NUS-WIDE.}
NUS-WIDE~\cite{chua2009nus} comprises of 269,648 images with a total of 81 visual concepts and 5,018 labels.
After filtering out unannotated samples, the training set and test set contain 125,449 and 83,898 images, respectively.
Notably, compared with other benchmarks, NUS-WIDE is more noisy and challenging.

\noindent\textbf{Evaluation Metrics.}
The mean average precision (mAP) is reported to evaluate the overall performance.
Following~\cite{wang2016cnn, chen2019multi, zhu2023scene}, we also report Class-wise Precision (CP), Recall (CR), F1 (CF1), and the average Overall Precision (OP), Recall (OR), F1 (OF1).
Note that ``CF1" and ``OF1" are more informative since Precision and Recall vary with the threshold.
To fairly compare with state-of-the-art, we further report top-3 results on MS-COCO and NUS-WIDE.

\begin{table*}[t]
    \centering
    \scalebox{0.94}{
        \small
        \begin{tabular}{p{58pt}<{\raggedright}p{43pt}<{\centering}p{43pt}<{\centering}|p{19pt}<{\centering}p{14pt}<{\centering}p{14pt}<{\centering}p{14pt}<{\centering}p{14pt}<{\centering}p{14pt}<{\centering}p{15pt}<{\centering}|p{14pt}<{\centering}p{14pt}<{\centering}p{14pt}<{\centering}p{14pt}<{\centering}p{14pt}<{\centering}p{15pt}<{\centering}}
        \toprule[1pt]
        \multirow{2}{*}{\hspace{-5pt}Method} & \multirow{2}{*}{Backbone} & \multirow{2}{*}{Resolution} & \multirow{2}{*}{mAP} & \multicolumn{6}{c|}{ALL} & \multicolumn{6}{c}{Top-3}\\
        & & & & CP & CR & CF1 & OP & OR & OF1 & CP & CR & CF1 & OP & OR & OF1 \\
        \bottomrule[0.5pt]
        \hspace{-5pt}ML-GCN~\cite{chen2019multi} & ResNet101 & (448, 448) & 83.0 & 85.1 & 72.0 & 78.0 & 85.8 & 75.4 & 80.3 & 89.2 & 64.1 & 74.6 & 90.5 & 66.5 & 76.7\\
        \hspace{-5pt}CMA~\cite{you2020cross} & ResNet101 & (448, 448) & 83.4 & 82.1 & 73.1 & 77.3 & 83.7 & 76.3 & 79.9 & 87.2 & 64.6 & 74.2 & 89.1 & 66.7 & 76.3\\
        \hspace{-5pt}TSGCN~\cite{xu2020joint} & ResNet101 & (448, 448) & 83.5 & 81.5 & 72.3 & 76.7 & 84.9 & 75.3 & 79.8 & 84.1 & 67.1 & 74.6 & 89.5 & 69.3 & 69.3\\
        \hspace{-5pt}CSRA~\cite{zhu2021residual} & ResNet101 & (448, 448) & 83.5 & 84.1 & 72.5 & 77.9 & 85.6 & 75.7 & 80.3 & 88.5 & 64.2 & 74.4 & 90.4 & 66.4 & 76.5\\
        \hspace{-5pt}ASL~\cite{ridnik2021asymmetric} & ResNet101 & (448, 448) & 85.0 & - & - & 80.3 & - & - & 82.3 & - & - & - & - & - & -\\
        \hspace{-5pt}TDRL~\cite{zhao2021transformer} & ResNet101 & (448, 448) & 84.6 & 86.0 & 73.1 & 79.0 & 86.6 & 76.4 & 81.2 & 89.9 & 64.4 & 75.0 & 91.2 & 67.0 & 77.2 \\
        \hspace{-5pt}Q2L-R101~\cite{liu2021query2label} & ResNet101 & (448, 448) & 84.9 & 84.8 & 74.5 & 79.3 & 86.6 & 76.9 & 81.5 & 78.0 & 69.1 & 73.3 & 80.7 & 70.6 & 75.4\\
        \hspace{-5pt}SALGL~\cite{zhu2023scene} & ResNet101 & (448, 448) & 85.8 & \textbf{87.2} & 74.5 & 80.4 & \textbf{87.8} & 77.6 & 82.4 & \textbf{90.4} & 65.7 & 76.1 & \textbf{91.9} & 67.9 & 78.1\\
        \rowcolor{Gray}\hspace{-5pt}\textbf{PVLR} & ResNet101 & (448, 448) & \textcolor{red}{\textbf{88.2}} & 82.2 & \textbf{82.2} & \textcolor{red}{\textbf{82.2}} & 82.8 & \textbf{85.4} & \textcolor{red}{\textbf{84.1}} & 88.5 & \textbf{69.0} & \textcolor{red}{\textbf{77.6}} & 90.3 & \textbf{71.2} & \textcolor{red}{\textbf{79.6}} \\
        \cline{1-16}
        \hspace{-5pt}\rule{0pt}{9pt}SSGRL~\cite{chen2019learning} & ResNet101 & (576, 576) & 83.6 & \textbf{89.5} & 68.3 & 76.9 & \textbf{91.2} & 70.7 & 79.3 & \textbf{91.9} & 62.1 & 73.0 & \textbf{93.6} & 64.2 & 76.0\\
        \hspace{-5pt}C-Tran~\cite{lanchantin2021general} & ResNet101 & (576, 576) & 85.1 & 86.3 & 74.3 & 79.9 & 87.7 & 76.5 & 81.7 & 90.1 & 65.7 & 76.0 & 92.1 & 71.4 & 77.6\\
        \hspace{-5pt}ADD-GCN~\cite{ye2020attention} & ResNet101 & (576, 576) & 85.2 & 84.7 & 75.9 & 80.1 & 84.9 & 79.4 & 82.0 & 88.8 & 66.2 & 75.8 & 90.3 & 68.5 & 77.9\\
        \hspace{-5pt}TDRL~\cite{zhao2021transformer} & ResNet101 & (576, 576) & 86.0 & 87.0 & 74.7 & 80.1 & 87.5 & 77.9 & 82.4 & 90.7 & 65.6 & 76.2 & 91.9 & 68.0 & 78.1 \\
        \hspace{-5pt}Q2L-R101~\cite{liu2021query2label} & ResNet101 & (576, 576) & 86.5 & 85.8 & 76.7 & 81.0 & 87.0 & 78.9 & 82.8 & 90.4 & 66.3 & 76.5 & 92.4 & 67.9 & 78.3\\
        \hspace{-5pt}SALGL~\cite{zhu2023scene} & ResNet101 & (576, 576) & 87.3 & 87.8 & 76.8 & 81.9 & 88.1 & 79.5 & 83.6 & 91.1 & 66.9 & 77.2 & 92.4 & 69.0 & 79.0 \\
        \rowcolor{Gray}\hspace{-5pt}\textbf{PVLR} & ResNet101 & (576, 576) & \textcolor{red}{\textbf{88.8}} & 82.7 & \textbf{83.1} & \textcolor{red}{\textbf{82.9}} & 83.1 & \textbf{86.1} & \textcolor{red}{\textbf{84.6}} & 88.6 & \textbf{69.6} & \textcolor{red}{\textbf{77.9}} & 90.6 & \textbf{71.5} & \textcolor{red}{\textbf{79.9}} \\
        \cline{1-16}
        \hspace{-5pt}\rule{0pt}{9pt}M3TR~\cite{zhao2021m3tr} & ViT-B/16 & (448, 448) & 87.5 & \textbf{88.4} & 77.2 & 82.5 & \textbf{88.3} & 79.8 & 83.8 & \textbf{91.9} & 68.1 & 78.2 & \textbf{92.6} & 69.6 & 79.4\\
        \hspace{-5pt}PatchCT~\cite{li2023patchct} & ViT-B/16 & (448, 448) & 88.3 & 83.3 & 82.3 & 82.6 & 84.2 & 83.7 & 83.8 & 90.7 & 69.7 & 78.8 & 90.3 & 70.8 & 79.8 \\
        \rowcolor{Gray}\hspace{-5pt}\textbf{PVLR} & ViT-B/16 & (448, 448) & \textcolor{red}{\textbf{90.5}} & 85.1 & \textbf{84.6} & \textcolor{red}{\textbf{84.9}} & 85.1 & \textbf{87.3} & \textcolor{red}{\textbf{86.2}} & 91.1 & \textbf{70.9} & \textcolor{red}{\textbf{79.7}} & 92.0 & \textbf{72.6} & \textcolor{red}{\textbf{81.2}}\\
        \bottomrule[1pt]
        \end{tabular}
    }
    \caption{\textbf{Comparison (\%) to state-of-the-art methods on MS-COCO.} Results with different backbone and input resolution are reported. Among them, mAP, OF1, and CF1 are the primary metrics (highlighted in \textcolor{red}{\textbf{red}}) as the others may be significantly affected by the threshold.}
    \label{tab:main_coco}
\end{table*}

\begin{table*}[t]
    \centering
    \scalebox{0.92}{
        \small
        \begin{tabular}{p{49pt}<{\raggedright}|p{10pt}<{\centering}p{10pt}<{\centering}p{10pt}<{\centering}p{10pt}<{\centering}p{11pt}<{\centering}p{10pt}<{\centering}p{10pt}<{\centering}p{10pt}<{\centering}p{10pt}<{\centering}p{10pt}<{\centering}p{10pt}<{\centering}p{10pt}<{\centering}p{12pt}<{\centering}p{12pt}<{\centering}p{13pt}<{\centering}p{10pt}<{\centering}p{10pt}<{\centering}p{10pt}<{\centering}p{10pt}<{\centering}p{12pt}<{\centering}|p{14pt}<{\centering}}
        \toprule[1pt]
        \hspace{-5pt}Method & aero & bike & bird & boat & bottle & bus & car & cat & chair & cow & table & dog & horse & motor & person & plant & sheep & sofa & train & tv & mAP\\
        \bottomrule[0.5pt]
        \hspace{-5pt}SSGRL~\cite{chen2019learning} & 99.5 & 97.1 & 97.6 & 97.8 & 82.6 & 94.8 & 96.7 & 98.1 & 78.0 & 97.0 & 85.6 & 97.8 & 98.3 & 96.4 & 98.8 & 84.9 & 96.5 & 79.8 & 98.4 & 92.8 & 93.4\\
        \hspace{-5pt}ML-GCN~\cite{chen2019multi} & 99.5 & 98.5 & \textbf{98.6} & 98.1 & 80.8 & 94.6 & 97.2 & 98.2 & 82.3 & 95.7 & 86.4 & 98.2 & 98.4 & 96.7 & 99.0 & 84.7 & 96.7 & 84.3 & 98.9 & 93.7 & 94.0\\
        \hspace{-5pt}TSGCN~\cite{xu2020joint} & 98.9 & 98.5 & 96.8 & 97.3 & 87.5 & 94.2 & 97.4 & 97.7 & 84.1 & 92.6 & 89.3 & 98.4 & 98.0 & 96.1 & 98.7 & 84.9 & 96.6 & 87.2 & 98.4 & 93.7 & 94.3\\
        \hspace{-5pt}ASL~\cite{ridnik2021asymmetric} & - & - & - & - & - & - & - & - & - & - & - & - & - & - & - & - & - & - & - & - & 94.4\\
        \hspace{-5pt}CSRA~\cite{zhu2021residual} & \textbf{99.9} & 98.4 & 98.1 & \textbf{98.9} & 82.2 & 95.3 & 97.8 & 97.9 & 84.6 & 94.8 & \textbf{90.8} & 98.1 & 97.6 & 96.2 & 99.1 & 86.4 & 95.9 & 88.3 & 98.9 & 94.4 & 94.7\\
        \hspace{-5pt}SALGL~\cite{zhu2023scene} & \textbf{99.9} & 98.8 & 98.3 & 98.2 & 81.6 & 96.5 & \textbf{98.1} & 97.8 & 85.2 & 97.0 & 89.6 & 98.5 & 98.7 & 97.1 & \textbf{99.2} & \textbf{86.9} & 96.4 & \textbf{89.9} & 99.5 & 95.2 & 95.1\\
        \rowcolor{Gray} \hspace{-5pt}\textbf{PVLR} & \textbf{99.9} & 98.1& 98.5& 98.8 & \textbf{88.8} & \textbf{98.7} & 97.1 & \textbf{99.4} & \textbf{86.3} & \textbf{98.6} & 87.4 & \textbf{99.3} & \textbf{98.8} & \textbf{98.5} & 98.8 & 85.7 & \textbf{99.4} & 86.2 & 98.7 & \textbf{96.0} & \textbf{95.5} \\
        \cline{1-22}
        \hspace{-5pt}\rule{0pt}{9pt}Q2L-TRL~\cite{liu2021query2label} & 99.9 & 98.9 & 99.0 & 98.4 & 87.7 & 98.6 & 98.8 & 99.1 & 84.5 & 98.3 & 89.2 & 99.2 & 99.2 & \textbf{99.2} & 99.3 & 90.2 & 98.8 & 88.3 & 99.5 & 95.5 & 96.1\\
        \hspace{-5pt}M3TR\dag~\cite{zhao2021m3tr} & 99.9 & 99.3 & 99.1 & 99.1 & 84.0 & 97.6 & 98.0 & 99.0 & 85.9 & \textbf{99.4} & 93.9 & \textbf{99.5} & 99.4 & 98.5 & 99.2 & 90.3 & 99.7 & 91.6 & 99.8 & 96.0 & 96.5\\
        \hspace{-5pt}PatchCT\dag~\cite{li2023patchct} & \makebox[16pt][r]{\textbf{100.0}} & \textbf{99.4} & 98.8 & 99.3 & 87.2 & 98.6 & 98.8 & 99.2 & 87.2 & 99.0 & \textbf{95.5} & 99.4 & \textbf{99.7} & 98.9 & 99.1 & 91.8 & 99.5 & \textbf{94.5} & \textbf{99.5} & 96.3 & 97.1\\
        \rowcolor{Gray} \hspace{-5pt}\textbf{PVLR}\dag & \makebox[16pt][r]{\textbf{100.0}} & 99.2& \textbf{99.3}& \textbf{99.4}& \textbf{91.1}& \textbf{99.8}& \textbf{99.0} & \textbf{99.6}& \textbf{90.6}&
        \textbf{99.4}& 93.0 & \textbf{99.5}& 99.4& 98.9& \textbf{99.5}& \textbf{92.1}& \makebox[16pt][r]{\textbf{100.0}} & 91.4&
        99.3& \textbf{97.4} &\textbf{97.4} \\
        \bottomrule[1pt]
        \end{tabular}
    }
    \caption{\textbf{Comparison (\%) to state-of-the-art methods on Pascal VOC 2007.} Results are reported in terms of class-wise average precision (AP) and mean average precision (mAP). \dag~indicates the ViT-B/16 backbone is used.}
    \label{tab:main_voc}
\end{table*}

\subsection{Implementation Details}
We use CLIP~\cite{radford2021learning} to extract textual embeddings and visual features.
Unless specified, the ResNet-101 is used as the image encoder.
The text encoder remains frozen in the training phase.
The number of learnable prompt tokens $L$ is set to $4$.
$\gamma^{+}$ and $\gamma^{-}$ are set as $0$ and $2$, respectively.
The hyper-parameter $\lambda$ is set as $4.0$.
The input images are resized to $448\times 448$ in both training and testing stages.
The network is trained for $30$ epochs using AdamW~\cite{loshchilov2017decoupled} optimizer with a batch size of $64$.
The learning rate is set as $0.0001$ and decays with the cosine policy.
Following previous works~\cite{ridnik2021asymmetric, zhu2023scene}, we apply exponential moving average to model parameters with a decay of 0.9997.

\subsection{Comparison with State-of-the-art}
The comparisons on MS-COCO, PASCAL VOC 2007, and NUS-WIDE are shown in Table~\ref{tab:main_coco}, Table~\ref{tab:main_voc} and Table~\ref{tab:main_nus}, respectively.
PVLR achieves state-of-the-art performance across various backbones and resolutions on all datasets, surpassing other methods with a decent margin.
On the MS-COCO, compared with SALGL~\cite{zhu2023scene} that utilized linguistic modality while hindering its role, PVLR exhibits considerable performance gains, exceeding them by 2.4\% mAP, suggesting the superiority of fully exploiting linguistic modality.
Compared with ML-GCN~\cite{chen2019multi} that also maps label representations into category centers while neglecting the visual context, PVLR achieves 5.2\% gains in mAP, demonstrating the effectiveness of learning input-adaptive category centers.
Moreover, our method outperforms all other methods on resolution of $576\times 576$ and ViT-B/16 backbone, surpassing previous state-of-the-art by 1.5\% and 2.2\% mAP respectively.
On the NUS-WIDE, our method surpasses all other methods on ResNet101 and ViT-B/16 backbones, achieving 67.4\% and 69.0\% mAP, respectively, which demonstrates the robustness of PVLR when addressing the noisy real-world images.
On the PASCAL VOC 2007, PVLR also outperforms all other methods.
With ViT-B/16 backbone, the AP of our method on all 20 categories exceeds 91.1\%,
which demonstrates the effectiveness of our method in handling objects of distinct sizes and semantics.
The experimental results clearly confirm the effectiveness of our proposed method, 
and also show good generalizability to different network architectures.

\begin{table}[t]
    \centering
    \scalebox{0.90}{
        \small
        \begin{tabular}{p{60pt}<{\raggedright}|p{25pt}<{\centering}p{20pt}<{\centering}p{20pt}<{\centering}|p{20pt}<{\centering}p{20pt}<{\centering}}
        \toprule[1pt]
        \multirow{2}{*}{\hspace{-4pt}Method} & \multirow{2}{*}{mAP} & \multicolumn{2}{c|}{ALL} & \multicolumn{2}{c}{Top-3}\\
         & & CF1 & OF1 & CF1 & OF1 \\
        \bottomrule[0.5pt]
        \hspace{-4pt}CMA~\cite{you2020cross} & 61.4 & 60.5 & 73.7 & 55.5 & 70.0\\
        \hspace{-4pt}GM-MLIC~\cite{wu2021gm} & 62.2 & 61.0 & 74.1 & 55.3 & \textbf{72.5}\\
        \hspace{-4pt}ICME~\cite{chen2019multi2} & 62.8 & 60.7 & 74.1 & 56.3 & 70.6\\
        \hspace{-4pt}ASL~\cite{ridnik2021asymmetric} & 63.9 & 62.7 & 74.6 & - & - \\
        \hspace{-4pt}SALGL~\cite{zhu2023scene} & 66.3 & 64.1 & 75.4 & 59.5 & 71.0\\
        \rowcolor{Gray} \hspace{-4pt}\textbf{PVLR} & \textbf{67.4} & \textbf{65.1} & \textbf{75.5} & \textbf{60.4} & 71.2 \\
        \bottomrule[0.5pt]
        \hspace{-4pt}Q2L-TRL~\cite{liu2021query2label} & 66.3 & 64.0 & 75.0 & - & - \\
        \hspace{-4pt}PatchCT\dag~\cite{li2023patchct} & 68.1 & 65.5 & 74.7 & 61.2 & 71.0\\
        \rowcolor{Gray} \hspace{-4pt}\textbf{PVLR}\dag & \textbf{69.0} & \textbf{65.6} & \textbf{76.0} & \textbf{62.0} & \textbf{71.7}\\
        \bottomrule[1pt]
        \end{tabular}
    }
    \caption{\textbf{Comparison (\%) to state-of-the-art methods on NUS-WIDE.} \dag~indicates ViT-B/16 backbone is used.}
    \label{tab:main_nus}
\end{table}
\begin{table}[t]
\hspace{-0.15cm}
    \centering
    \scalebox{0.88}{
        \small
        \begin{tabular}{p{14pt}<{\centering}p{14pt}<{\centering}p{14pt}<{\centering}p{14pt}<{\raggedright}|p{16pt}<{\centering}p{14pt}<{\centering}p{15pt}<{\centering}|p{16pt}<{\centering}p{14pt}<{\centering}p{15pt}<{\centering}}
        \toprule[1pt]
        \multirow{2}{*}{KAP} & \multirow{2}{*}{CAP} & \multirow{2}{*}{IFM} & \multirow{2}{*}{\hspace{-4pt}DMA} & \multicolumn{3}{c|}{MS-COCO} & \multicolumn{3}{c}{NUS-WIDE} \\
        &&&& mAP & CF1 & OF1 & mAP & CF1 & OF1 \\
        \bottomrule[0.5pt]
        &&&& 81.6 & 54.6 & 47.7 & 58.6 & 29.1 & 36.6 \\
        \ding{51} & & & & 84.1 & 78.7 & 82.1 & 62.6 & 57.8 & 74.8 \\
        & \ding{51} & & & 86.9 & 80.4 & 82.9 & 65.9 & 64.0 & 74.2 \\
        \ding{51} & \ding{51} & \ding{51} & & 87.5 & 82.0 & 83.9 & 66.7 & 65.2 & 75.1 \\
        \rowcolor{Gray} \ding{51} & \ding{51} & \ding{51} & \ding{51} & \textbf{88.2} & \textbf{82.2} & \textbf{84.1} & \textbf{67.4} & \textbf{65.1} & \textbf{75.5}\\
        \bottomrule[1pt]
        \end{tabular}
    }
    \caption{\textbf{Ablation study (\%) on the proposed modules.} The baseline method in the first row utilizes the pure category names to extract label representations.}
    \label{tab:abl1}
\end{table}
\subsection{Ablation Studies}

\textbf{Effect of proposed modules.}
The foremost thing we are interested in is the improvements brought by the proposed KAP, CAP, IFM and DMA.
To verify this, we set a baseline method that utilizes pure category names to extract label representations and no further interactions are performed between modalities.
As shown in Table~\ref{tab:abl1}, the performance of using pure label names is unsatisfactory, leading to poor CF1 and OF1. 
While KAP heals the performance significantly.
We attribute this to that KAP effectively extracts knowledge from the language model. 
Additionally, applying CAP is extremely effective, improving mAP by 5.3\% and 7.3\% on MS-COCO and NUS-WIDE respectively. 
Moreover, employing IFM and DMA can also steadily improve the performance, suggesting that aggregating such static and dynamic information is helpful under the variable visual scenes.
These observations fairly verify the effectiveness of our proposed four modules.

\noindent\textbf{Is using label representations as category centers better?}
In this sub-section, we verify the effectiveness of using 
label representations as category centers.
Besides our proposed method, we set a ``Classifier Learning'' experiment as reference,
where we we perform a one-way interaction and $C$ traditional classifiers are learned (as shown in Figure~\ref{fig:intro}). 
Accordingly, ``Label Rep.'' denotes the category centers are mapped from label representations, and ``Label Rep. + DMA'' indicates that the DMA is further employed. 
As shown in Table~\ref{tab:abl2}, the low performance of ``Label Rep.'' indicates that generating category centers solely from \textit{static} label representations is inferior, 
while this can be improved by constructing \textit{dynamic} category centers as ``Label Rep. + DMA''.
Moreover, PVLR surpasses all three reference methods,
which further demonstrates the superiority of our proposed components including KAP, CAP, IFM and DMA.
Similar conclusions can be drawn when using BERT~\cite{devlin2018bert} as the language model (shown in Table~\ref{tab:abl2}).

\begin{table}[t]
    \centering
    \scalebox{0.84}{
        \small
        \begin{tabular}{p{66pt}<{\raggedright}|p{39pt}<{\centering}|p{13pt}<{\centering}p{13pt}<{\centering}p{13pt}<{\centering}|p{13pt}<{\centering}p{13pt}<{\centering}p{13pt}<{\centering}}
        \toprule[1pt]
        \hspace{-5pt}\multirow{2}{*}{Method} & \multirow{2}{*}{Text} & \multicolumn{3}{c|}{MS-COCO} & \multicolumn{3}{c}{NUS-WIDE} \\
        && mAP & CF1 & OF1 & mAP & CF1 & OF1 \\
        \bottomrule[0.5pt]
        \hspace{-5pt}Classifier Learning & CLIP-Text & 85.0 & 78.7 & 82.4 & 65.4 & 63.3 & 73.3 \\
        \hspace{-5pt}Label Rep. & CLIP-Text & 81.6 & 54.6 & 47.7 & 58.6 & 29.1 & 36.6  \\
        \hspace{-5pt}Label Rep. + DMA & CLIP-Text & 87.1 & 81.3 & 84.1 & 65.6 & 62.6 & 74.0 \\
        \rowcolor{Gray} & & \textbf{88.2} & \textbf{82.2} & \textbf{84.1} & \textbf{67.4} & \textbf{65.1} & \textbf{75.5} \\
            \rowcolor{Gray} \multirow{-2}{*}{\hspace{-5pt}PVLR} & \multirow{-2}{*}{CLIP-Text} & \makebox[19pt][r]{\textcolor[RGB]{50,205,50}{(\textbf{$\uparrow$3.2})}} & \makebox[19pt][r]{\textcolor[RGB]{50,205,50}{(\textbf{$\uparrow$3.5})}} & \makebox[19pt][r]{\textcolor[RGB]{50,205,50}{(\textbf{$\uparrow$1.7})}} & \makebox[19pt][r]{\textcolor[RGB]{50,205,50}{(\textbf{$\uparrow$2.0})}} & \makebox[19pt][r]{\textcolor[RGB]{50,205,50}{(\textbf{$\uparrow$1.8})}} & \makebox[19pt][r]{\textcolor[RGB]{50,205,50}{(\textbf{$\uparrow$2.2})}} \\
        \bottomrule[0.5pt]
        \hspace{-5pt}Classifier Learning & BERT$_{\text{Base}}$ & 84.8 & 78.3 & 82.1 & 64.8 & 63.1 & 73.7 \\
        \hspace{-5pt}Label Rep. & BERT$_{\text{Base}}$ & 83.6 & 78.1 & 81.8 & 60.2 & 57.5 & 73.6  \\
        \hspace{-5pt}Label Rep. + DMA & BERT$_{\text{Base}}$ & 87.3 & 81.0 & 83.6 & 65.9 & 62.4 & 74.8 \\
        \rowcolor{Gray} & & \textbf{88.0} & \textbf{82.2} & \textbf{83.9} & \textbf{66.9} & \textbf{64.7} & \textbf{74.9} \\
            \rowcolor{Gray} \multirow{-2}{*}{\hspace{-5pt}PVLR} & \multirow{-2}{*}{BERT$_{\text{Base}}$} & \makebox[19pt][r]{\textcolor[RGB]{50,205,50}{(\textbf{$\uparrow$3.2})}} & \makebox[19pt][r]{\textcolor[RGB]{50,205,50}{(\textbf{$\uparrow$3.9})}} & \makebox[19pt][r]{\textcolor[RGB]{50,205,50}{(\textbf{$\uparrow$1.8})}} & \makebox[19pt][r]{\textcolor[RGB]{50,205,50}{(\textbf{$\uparrow$2.1})}} & \makebox[19pt][r]{\textcolor[RGB]{50,205,50}{(\textbf{$\uparrow$1.6})}} & \makebox[19pt][r]{\textcolor[RGB]{50,205,50}{(\textbf{$\uparrow$1.2})}} \\
        \bottomrule[1pt]
        \end{tabular}
    }
    \caption{\textbf{Ablation study (\%) on the generation of category centers.} ``Label Rep.'' means the category centers are generated from label representations.}
    \label{tab:abl2}
\end{table}
\begin{figure*}[t]
    \centering
    \includegraphics[width=0.99\linewidth]{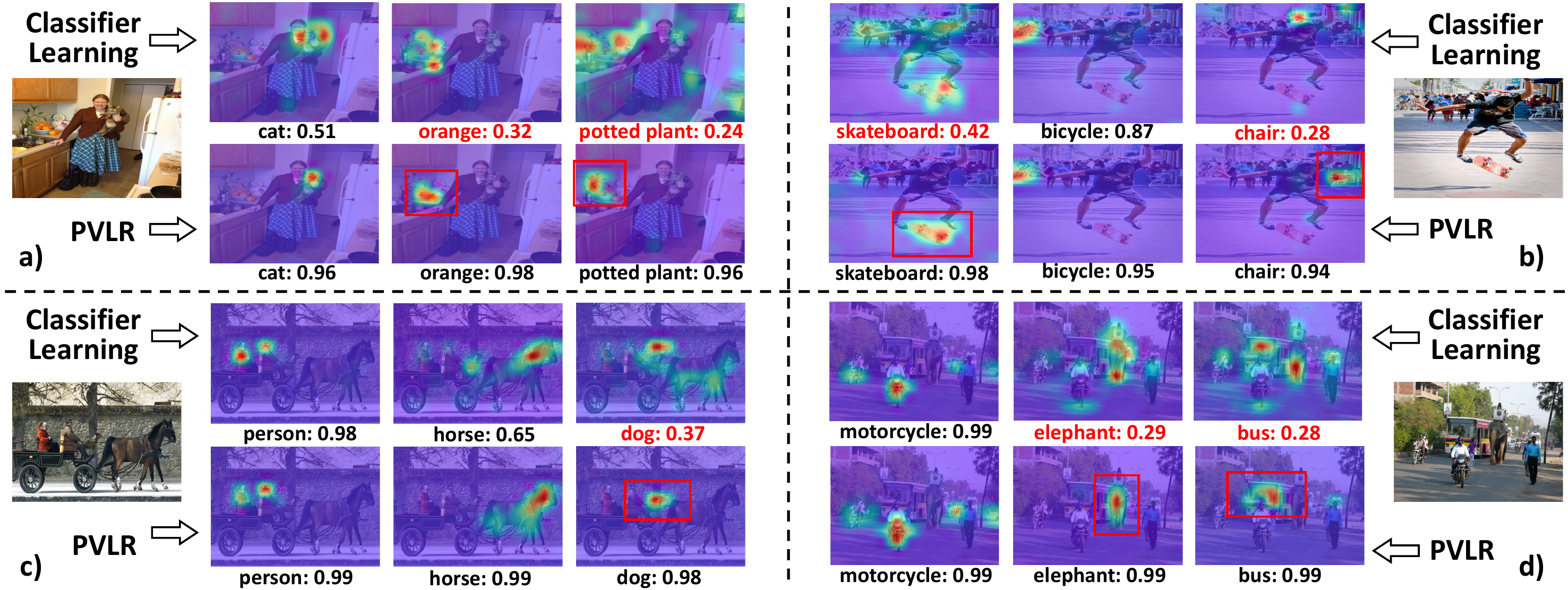}
    \caption{\textbf{Visualization of ``Classifier Learning'' approach and our proposed PVLR. }
    We present several labels for demonstration and those scores highlighted in \textcolor{red}{\textbf{red}} are not successfully recognized.
    PVLR can accurately perceive and localize small objects in a) and b), similar objects in c), and target objects in even complex context in d).}
	\label{fig:vis}
\end{figure*}

\noindent\textbf{Effect of prompting strategy.}
In Table~\ref{tab:abl3}, we explore different prompting strategies.
Specifically, using hard prompts in CAP yields degraded performance.
Conditioning the soft prompts on visual features before the text encoder leads to much more training time and inferior performance (a detailed explanation is provided in the \textbf{supplementary material}).
Therefore, interacting soft prompts with visual context after the text encoder is a better choice.
Moreover, using more prompt tokens (e.g., $L=8$ or $L=12$) yields better results on MS-COCO but inferior results on the noisy NUS-WIDE.
Therefore, we suggest using $4$ prompt tokens in CAP as a good trade-off.

\begin{table}[t]
    \centering
    \scalebox{0.87}{
        \small
        \begin{tabular}{p{79pt}<{\raggedright}|p{10pt}<{\centering}|p{14pt}<{\centering}p{14pt}<{\centering}p{14pt}<{\centering}|p{14pt}<{\centering}p{14pt}<{\centering}p{14pt}<{\centering}}
        \toprule[1pt]
        \hspace{-5pt}\multirow{2}{*}{Method} & \multirow{2}{*}{$L$} & \multicolumn{3}{c|}{MS-COCO} & \multicolumn{3}{c}{NUS-WIDE} \\
        & & mAP & CF1 & OF1 & mAP & CF1 & OF1 \\
        \bottomrule[0.5pt]
        \hspace{-5pt}CAP w/ hard prompts & - & 87.8 & 81.7 & 83.3 & 66.4 & 64.2 & 75.0 \\
        \hspace{-5pt}CAP w/ pre-interaction & 4 & 87.6 & 81.9 & 83.9 & 65.8 & 63.7 & 74.7 \\
        \hspace{-5pt}CAP w/ soft prompts & 4 & 88.2 & 82.2 & \textbf{84.1} & \textbf{67.4} & \textbf{65.1} & \textbf{75.5} \\
        \hspace{-5pt}CAP w/ soft prompts & 8 & 88.2 & \textbf{82.4} & 84.0 & 67.0 & 64.8 & 75.1 \\
        \hspace{-5pt}CAP w/ soft prompts & 12 & \textbf{88.4} & 82.3 & 84.0 & 66.9 & 64.6 & 75.0 \\
        \bottomrule[1pt]
        \end{tabular}
    }
    \caption{\textbf{Ablation study (\%) on the prompting strategy.} $L$ is the number of learnable prompt tokens.}
    \label{tab:abl3}
\end{table}

\begin{figure}[t]
    \centering
    \includegraphics[width=0.97\linewidth]{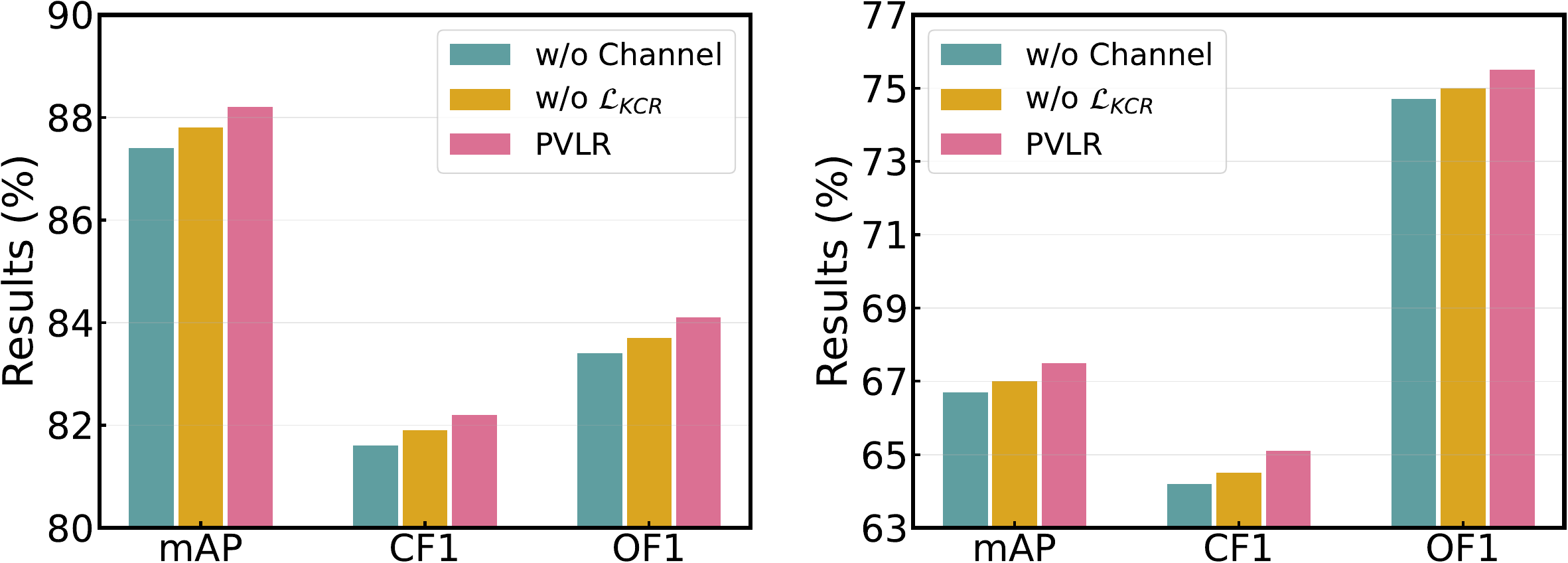}
 \caption{\textbf{Ablation study (\%) on the IFM.} 
 Results are reported on MS-COCO (left) and NUS-WIDE (right).}
	\label{fig:ifm}
\end{figure}

\noindent\textbf{Components of IFM.}
As shown in Figure~\ref{fig:ifm}, both $\mathcal{L}_{KCR}$ and the channel interaction improve the overall performance, while channel interaction is more important.
This indicates that the explicit channel-by-channel interaction is more effective in aggregating the dual prompting branches.

\subsection{Further Analyses}
\begin{figure}[t]
    \centering
    \includegraphics[width=0.97\linewidth]{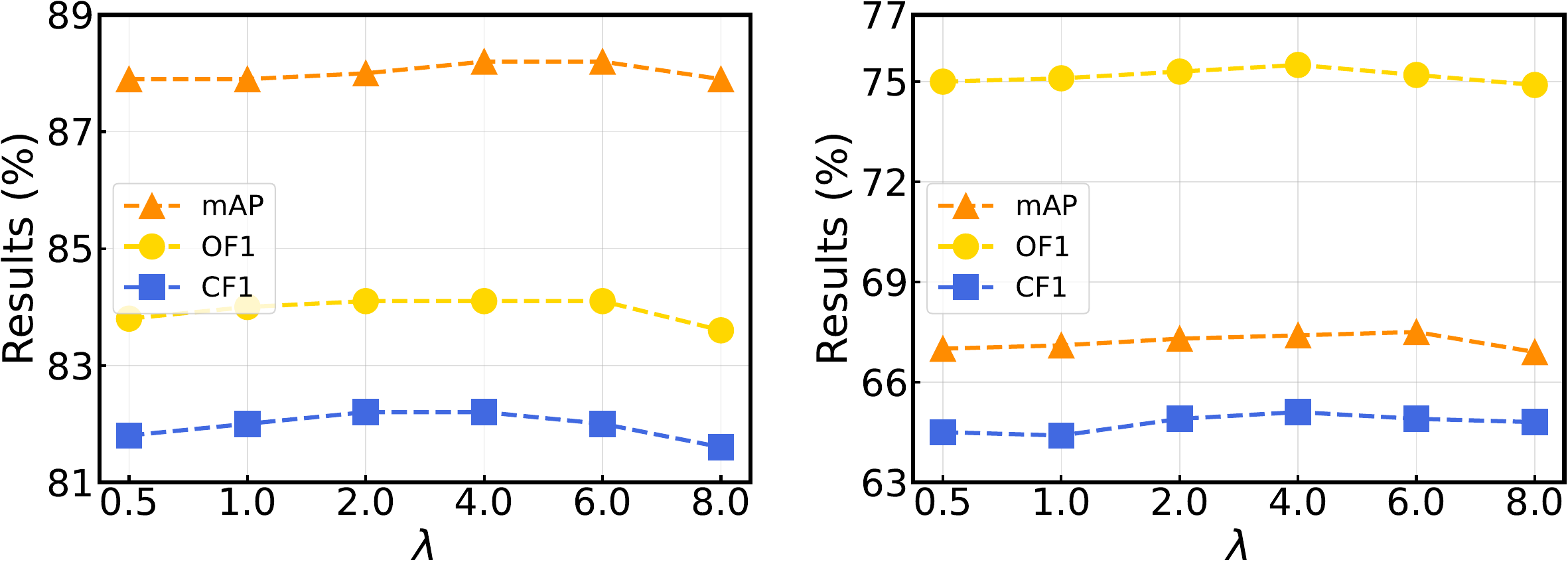}
 \caption{\textbf{Sensitivity analysis (\%) of $\lambda$.} Results are reported on MS-COCO (left) and NUS-WIDE (right).}
	\label{fig:lambda}
\end{figure}

\textbf{Sensitivity analysis of $\lambda$.}
As shown in Figure~\ref{fig:lambda}, we evaluate the parameter sensitivity of $\lambda$ in Eq.~\ref{eq:loss}. 
The results suggest that the performance of PVLR is generally stable,
while $\lambda=4$ obtains better performance.
This indicates the robustness of our method to $\lambda$.

\noindent\textbf{Visualization.}
In Figure~\ref{fig:vis}, e visualize the cross-attention map of the ``Classifier Learning'' approach and our proposed PVLR.
1) PVLR can more accurately localize small objects, e.g. \textit{orange} in a).
2) PVLR has the ability to distinguish similar objects, e.g. \textit{dog} and \textit{horse} in c) while Classifier Learning confuses them.
3) Classifier Learning exhibits poor capabilities in complex contexts such as d), while PVLR can still precisely localize the target areas.
More examples are provided in \textbf{supplementary material}.

\section{Conclusion}
\label{sec:con}
In this work, we propose PVLR, a novel visual-linguistic representation learning framework for multi-label image recognition.
To address the defects of existing multi-modal approaches, we propose four modules, namely KAP, CAP, IFM and DMA to fully exploit the linguistic modality and learn the context-aware label representations and semantic-related visual representations concurrently.
Extensive experiments show that PVLR achieves state-of-the-art performance on three widely used benchmarks.

\noindent\textbf{Limitations and broader impacts.}
One limitation is that the semantic knowledge extracted by the pre-trained vision-language model relies on the model's pretraining data.
This may introduce some unexpected noises to our method.
Additionally, in our training datasets, there are unannotated objects present in the images, which could impact the model's performance in real-world scenarios. 
Our aim in this paper is to develop a general method for multi-label image recognition without targeting specific applications, which does not directly involve specific societal issues.

{
    \small
    \bibliographystyle{ieeenat_fullname}
    \bibliography{main}
}

\clearpage
\appendix

\centerline{\textbf{\Large Appendix}}

\begin{table}[h]
    \centering
    \scalebox{0.96}{
        \small
        \begin{tabular}{p{53pt}<{\raggedright}p{56pt}<{\centering}p{43pt}<{\centering}p{45pt}<{\centering}}
        \toprule[1pt]
        \multirow{2}{*}{\hspace{-5pt}Strategy} & Training Speed & \multirow{2}{*}{MS-COCO} & \multirow{2}{*}{NUS-WIDE} \\
        & (seconds/batch) && \\
        \bottomrule[0.5pt]
        \hspace{-5pt}Pre-interaction & 2.40 & 87.6 & 65.8 \\
        \hspace{-5pt}Post-interaction & \textbf{0.41} & \textbf{88.2} & \textbf{67.4} \\
        \bottomrule[1pt]
        \end{tabular}
    }
    \caption{\textbf{Comparison on different strategies in CAP.} The training speed is reported as the training time of a complete forward pass and a backward pass, using a batch size of 64.}
    \label{tab:cap}
\end{table}
\begin{figure}[h]
    \centering
    \includegraphics[width=0.94\linewidth]{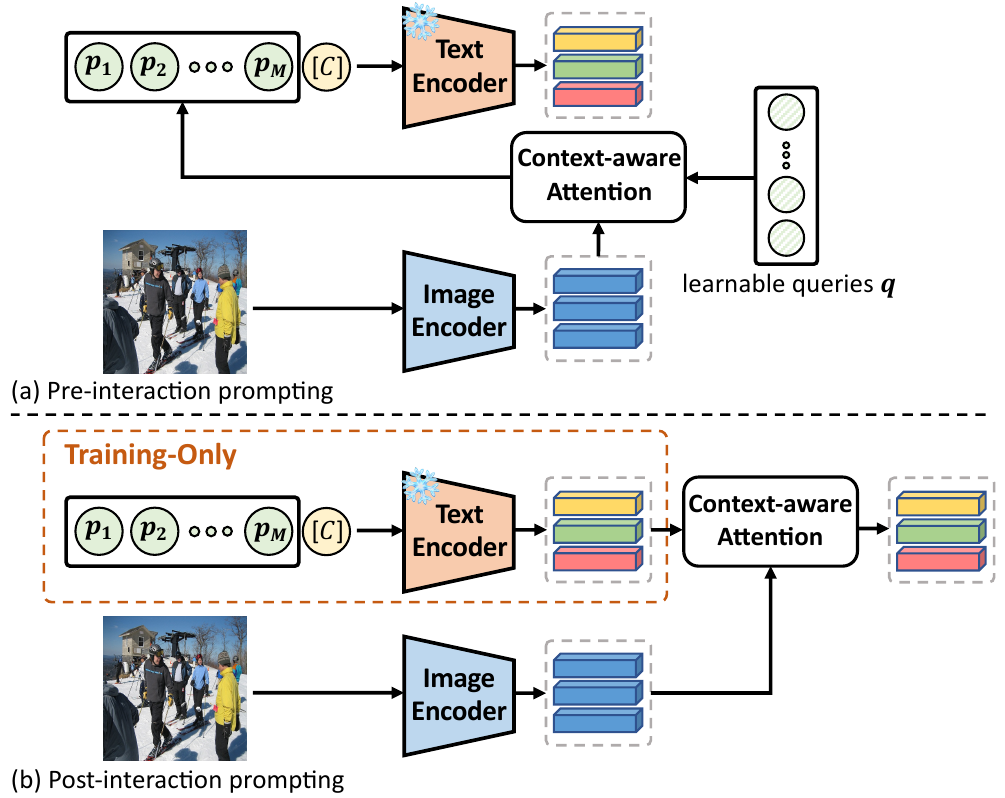}
    \caption{\textbf{Different prompting strategies in CAP.}}
	\label{fig:cap} 
\end{figure}

\section{Discussion on Prompting Strategies in CAP}
In our proposed CAP, the extracted label embeddings interact with visual features to capture context-aware label semantics.
In practice, there are two different strategies to condition the CAP on visual context.
As shown in Figure~\ref{fig:cap}, the first is ``\textit{pre-interaction prompting}'', where the interaction happens before the text encoder and thus the prompt tokens $\{\boldsymbol{p}_l\}_{l=1}^{L}$ are generated from the learnable queries and visual features.
The second is ``\textit{post-interaction prompting}'', where the interaction happens after the text encoder and the label embeddings are refined by the visual features.
In this paper, we choose the latter scheme since the former strategy requires an independent forward pass of instance-wise prompts through the text encoder, which consumes significant GPU memory and is less efficient than the latter, while the latter is a lightweight choice and exhibits better performance as reported in Table~\ref{tab:cap}.

\begin{table}[t]
    \centering
    \scalebox{0.96}{
        \small
        \begin{tabular}{p{30pt}<{\centering}p{30pt}<{\centering}|p{46pt}<{\centering}p{46pt}<{\centering}}
        \toprule[1pt]
        \hspace{-5pt}V-to-S & S-to-V & MS-COCO & NUS-WIDE \\
        \bottomrule[0.5pt]
        & & 87.5 & 66.7\\
        \ding{51} &  & 87.8 & 67.0 \\
        & \ding{51} & 87.7 & 66.2 \\
        \ding{51} & \ding{51} & \textbf{88.2} & \textbf{67.4} \\
        \bottomrule[1pt]
        \end{tabular}
    }
    \caption{\textbf{Ablation study (\%) on the aggregation of visual clues.}
    ``V-to-S'' and ``S-to-V'' denote visual-to-semantic attention and semantic-to-visual attention, respectively.}
    \label{tab:bid}
\end{table}

\section{Discussion on the Bidirectional Interaction}
One major distinction of our approach from previous methods is the bidirectional interaction between visual and linguistic modality.
In this section, we investigate the effectiveness of the bidirectional interaction.
Specifically, our proposed DMA consists of semantic-to-visual attention and visual-to-semantic attention.
By employing only one of them, we perform unidirectional interaction.
As shown in Table~\ref{tab:bid}, visual-to-semantic attention steadily improves the mAP, while semantic-to-visual attention hurts the performance on the noisy NUS-WIDE.
However, employing both jointly yields significant enhancement, indicating the superiority of bidirectional interaction.

\section{Discussion on the Roles of Visual Clues}
In PVLR, one core idea is to aggregate visual clues into label representations to yield input-adaptive category centers.
In this section, we investigate the effectiveness of such operations.
Specifically, we incorporate the visual clues in two aspects.
Firstly, we implicitly inject visual information into category centers through CAP.
Secondly, we explicitly incorporate visual information into category centers through DMA, which enhances the generalization.
The performance is shown in Table~\ref{tab:visual} by removing either of the two aggregations.
Compared to baseline, both implicit and explicit aggregation of visual information improve the mAP significantly, while employing them jointly achieves better results.

\begin{table}[t]
    \centering
    \scalebox{0.96}{
        \small
        \begin{tabular}{p{30pt}<{\centering}p{30pt}<{\centering}|p{46pt}<{\centering}p{46pt}<{\centering}}
        \toprule[1pt]
        \hspace{-5pt}Implicit & Explicit & MS-COCO & NUS-WIDE \\
        \bottomrule[0.5pt]
        & & 86.1 & 64.8 \\
        \ding{51} &  & 87.5 & 66.7 \\
        & \ding{51} & 87.9 & 66.9 \\
        \ding{51} & \ding{51} & \textbf{88.2} & \textbf{67.4} \\
        \bottomrule[1pt]
        \end{tabular}
    }
    \caption{\textbf{Ablation study (\%) on the aggregation of visual clues.} 
    PVLR implicitly injects visual information into category centers through CAP, while explicitly incorporating visual information into category centers through DMA.
    }
    \label{tab:visual}
\end{table}

\section{Additional Results with Different Backbones}
We examine our methods on more backbones in Table~\ref{tab:backbone}, including ResNet50 and ViT-B/32.
Notably, when using a patch size of 32 with ViT-B, the performance is inferior to ResNet101. This may be attributed to that the recognition of some small objects in multi-label images requires more fine-grained visual features.
Overall, PVLR achieves steady improvements with more powerful backbones, indicating the generalizability to network architectures.

\section{Qualitative Results}
\begin{figure*}[t]
    \centering
    \includegraphics[width=0.95\linewidth]{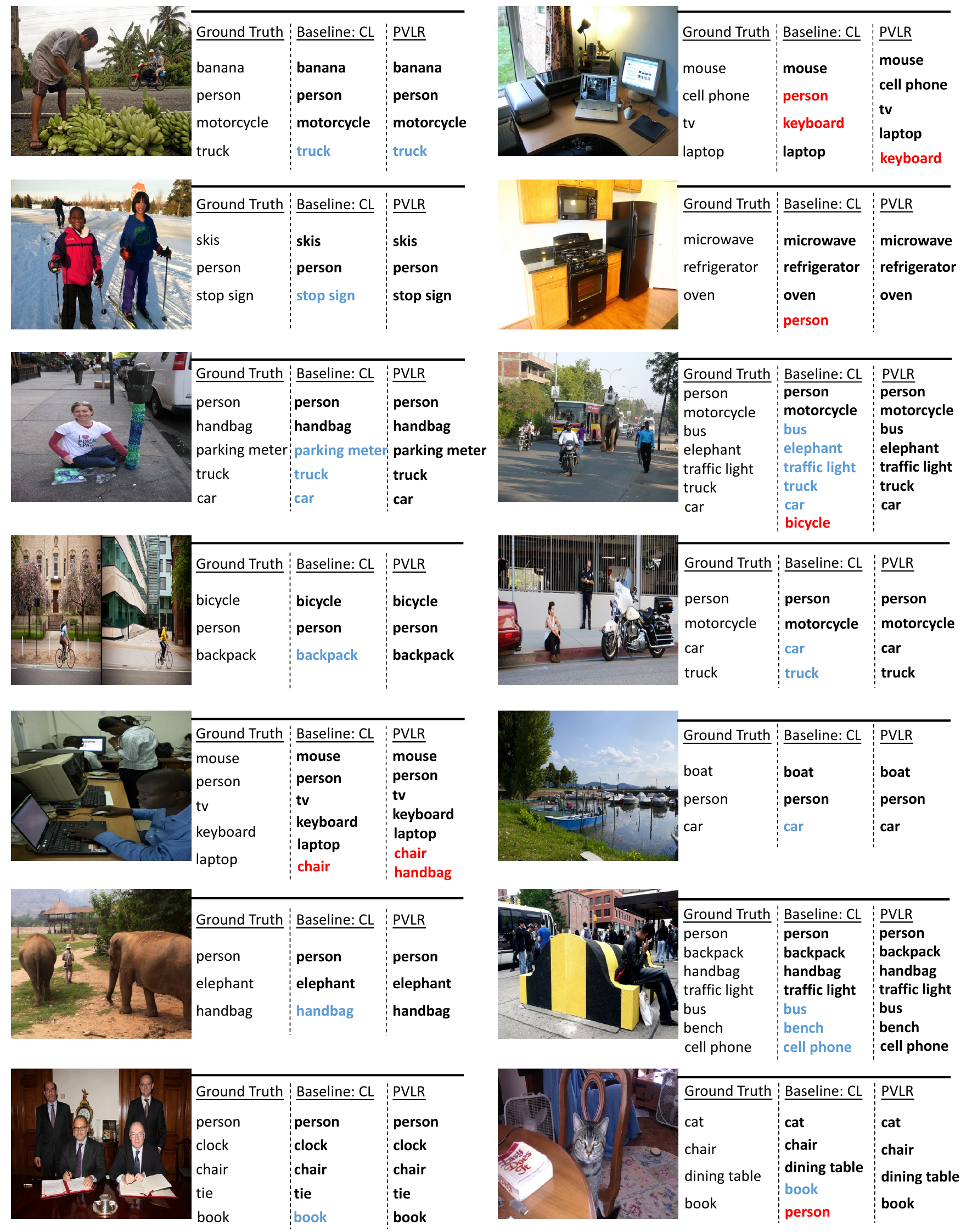}
 \caption{\textbf{Predictions of ``Classifier Learning'' (CL) baseline and our proposed PVLR.}
 Each sample is marked with three columns. From the left to right are ground truth labels, predictions of baseline and predictions of PVLR, respectively.
 Correct predictions are marked in \textbf{bold} font.
 Missed objects are highlighted in \textcolor[RGB]{93,173,226}{\textbf{blue}} and incorrect predictions are marked in \textcolor{red}{\textbf{red}}.
 Best viewed in colors.
 }
	\label{fig:pred}
\end{figure*}

In Figure~\ref{fig:pred}, we provide the predicted results of ``Classifier Learning'' approach and our proposed PVLR with a threshold of 0.5.
\textbf{1)} Baseline produces many false negative predictions while PVLR mitigates the issue.
\textbf{2)} Baseline tends to predict ``person'' in all scenarios, while always neglecting ``book'' as shown in the last row of Figure~\ref{fig:pred}.
We attribute this to that the static category centers in Baseline tend to overfit the distribution of the training set, resulting in high confidence in the frequently occurred labels such as ``person''. 
In contrast, dynamically constructing the context-aware category centers can alleviate this issue.

\begin{table}[t]
    \centering
    \scalebox{0.94}{
        \small
        \begin{tabular}{p{20pt}<{\raggedright}p{40pt}<{\centering}|p{15pt}<{\centering}p{15pt}<{\centering}p{15pt}<{\centering}|p{15pt}<{\centering}p{15pt}<{\centering}p{15pt}<{\centering}}
        \toprule[1pt]
        \multirow{2}{*}{\hspace{-5pt}Method} & \multirow{2}{*}{Backbone} & \multicolumn{3}{c|}{MS-COCO} & \multicolumn{3}{c}{NUS-WIDE} \\
        && mAP & CF1 & OF1 & mAP & CF1 & OF1 \\
        \bottomrule[0.5pt]
        \hspace{-5pt}PVLR & ResNet50 & 86.3 & 80.4 & 82.5 & 66.2 & 63.8 & 74.9\\
        \hspace{-5pt}PVLR & ResNet101 & 88.2 & 82.2 & 84.1 & 67.4 & 65.1 & 75.5 \\
        \hspace{-5pt}PVLR & ViT-B/32 & 86.9 & 80.9 & 82.6 & 67.1 & 64.8 & 75.5 \\
        \hspace{-5pt}PVLR & ViT-B/16 & 90.5 & 84.9 & 86.2 & 69.0 & 65.6 & 76.0 \\
        \bottomrule[1pt]
        \end{tabular}
    }
    \caption{\textbf{Additional results (\%) with different backbones.}}
    \label{tab:backbone}
\end{table}

\section{More Visualization Results}
\begin{figure*}[t]
    \centering
    \includegraphics[width=0.84\linewidth]{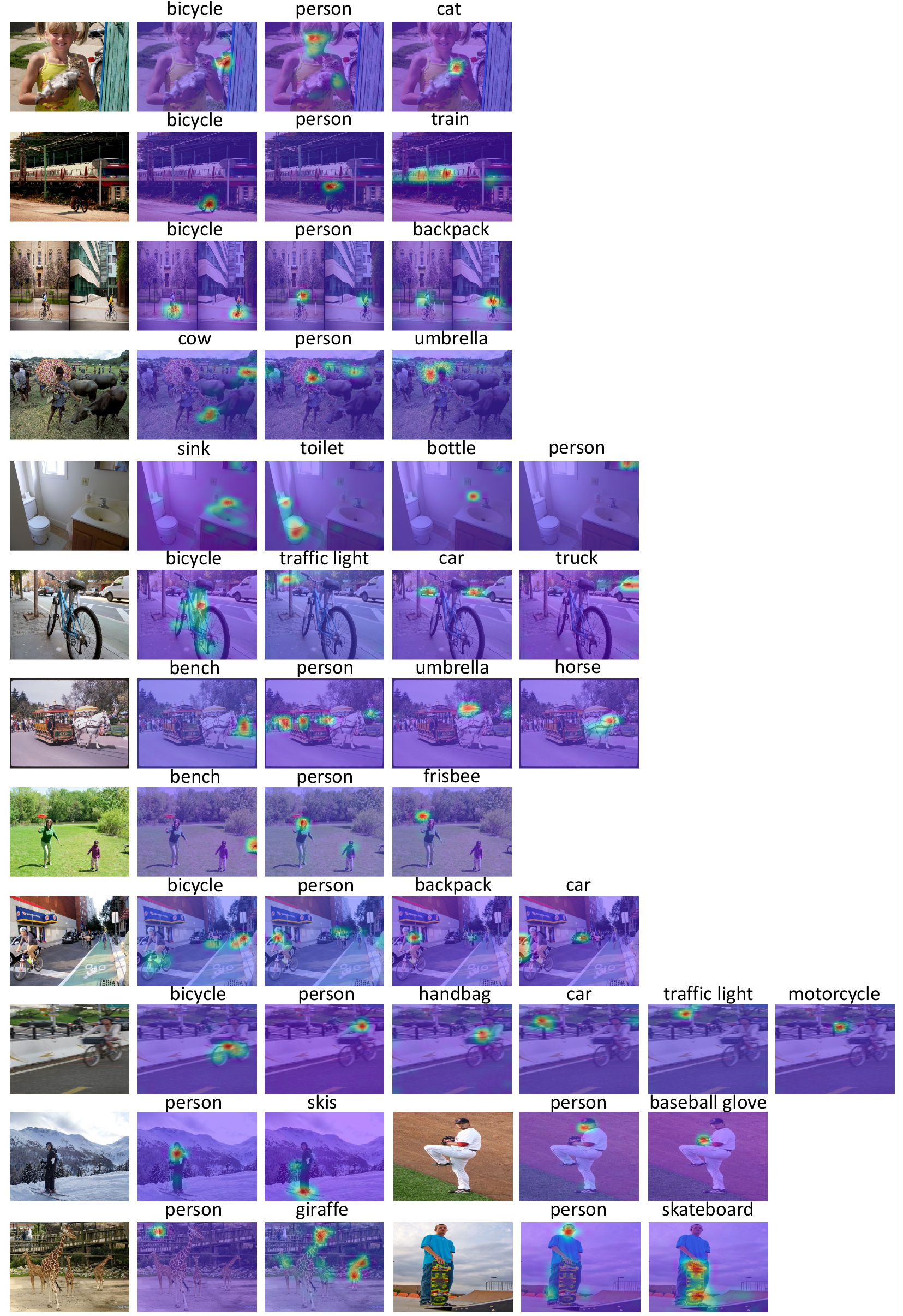}
 \caption{\textbf{More visualizations of cross-attention map.}
 Texts above the images denote the ground truth labels for the images.
 The brighter the color, the higher the attention to the corresponding area.
 Best viewed in colors.
 }
	\label{fig:cattnmap}
\end{figure*}

In Figure~\ref{fig:cattnmap}, we provide more visualization results of cross-attention maps on MS-COCO.
PVLR can accurately perceive and localize objects of distinct sizes and semantics, demonstrating the effectiveness of our method.

\end{document}